\documentclass[submission,copyright,creativecommons]{eptcs}
\providecommand{\event}{ICLP 2024} 

\usepackage{hyperref}
\usepackage{mathtools}
\usepackage{txfonts}
\usepackage{xspace}
\usepackage{underscore}   
\usepackage[T1]{fontenc}   

\newcommand{\bnf}  { \coloneqq }
\renewcommand{\phi}  { \varphi }
\newcommand{\eqdef}  { \stackrel{\mathsf{def}}{=} }
\newcommand{\lang}[1]  { {\cal L}_{#1} }
\newcommand{\tuple}[1]  { \langle #1 \rangle }
\newcommand{\classnot}  { {\sim} } 
\newcommand{\Prop}  { \mathbb{P} }

\newcommand\restr[2]  { \ensuremath{\left.#1\right|_{#2}} }

\newcommand{\suchthat}  { \ : \ }
\newcommand{\onequote}[1]  {`#1'}
\newcommand{\doublequotes}[1]  {``#1''}
\newcommand{\set}[1]  { \{ #1 \} }
\newcommand{\bigset}[1]  { \big\{ #1 \big\} }

\newcommand{\logic}[1]  { \ensuremath{\mathsf{#1}} }
\newcommand{\sfive}  { \logic{S5} }
\newcommand{\KD}  { \logic{KD45} }
\newcommand{\SW}  { \logic{SW5} }
\newcommand{\HT}  { \logic{HT} }
\newcommand{\EHT}  { \logic{EHT} }
\newcommand{\EHTyirmib}  { \logic{EHT_{\!\!\scriptscriptstyle{20}} } }
\newcommand{\ASP}  { \logic{ASP} }
\newcommand{\EL}  { \logic{EL} }
\newcommand{\ES}  { \logic{ES} }
\newcommand{\EASP}  { \logic{EASP} }
\newcommand{\EEL}  { \logic{EEL} }
\newcommand{\AEEL}  { \logic{AEEL} }
\newcommand{\RAEEL}  { \logic{RAEEL} }
\newcommand{\FAEEL}  { \logic{FAEEL} }
\newcommand{\AI}  { \logic{AI} }
\newcommand{\ELP}  { \logic{ELP} }

\newcommand{\naf}  { \texttt{NAF} }
\newcommand{\lpnot}  { \mathtt{not} \, }
\newcommand{\notlp}  { \mathtt{not} }
\newcommand{\anset}  { \texttt{AS} }
\newcommand{\Body}  { \mathtt{body} }
\newcommand{\intimp}  { \rightarrow }
\newcommand{\intnot}  { \neg }
\newcommand{\trOf}[1]  { #1 ^* }

\newcommand{\epispec}  { \Pi }
\newcommand{\K}  { \mathsf{K} }
\newcommand{\M}  { \mathsf{M} }
\newcommand{\Khat}  { \hat{\mathsf{K}} }

\newcommand{\ESdoksandort}  { \logic{ES_{\scriptscriptstyle{94}} } }
\newcommand{\ESonbir}  { \logic{ES_{\scriptscriptstyle{11}} } }
\newcommand{\ESyirmia}  { \logic{ES_{\scriptscriptstyle{20a}} } }
\newcommand{\ESyirmib}  { \logic{ES_{\scriptscriptstyle{20b}} } }
\newcommand{\ESyirmibir}  { \logic{ES_{\scriptscriptstyle{21}} } }

\newcommand{\there}  { \mathcal{T} }
\newcommand{\weak}  { \texttt{s} }
\newcommand{\functional}  { \texttt{f} }
\newcommand{\relational}  { \texttt{r} }
\newcommand{\weakrelational}  { \weak_{\!\relational} }
\newcommand{\tfunctional}  { \texttt{t}_{\!\functional} }
\newcommand{\trelational}  { \texttt{t}_{\!\relational} }
\newcommand{\EEM} { \texttt{EEM} }
\newcommand{\AEEM}  { \texttt{AEEM} }
\newcommand{\EEMs}  { \texttt{EEMs} }
\newcommand{\AEEMs}  { \texttt{AEEMs} }
\newcommand{\Ehtmodels}  { \models_{\scriptscriptstyle \EHT} }
\newcommand{\notEhtmodels}  { \not\models_{\scriptscriptstyle \EHT} }
\newcommand{\negof}[1]  { \widetilde{#1} }
\newcommand{\starmodels}  { \models^* }

\newcommand{\starmodelsrelational}  { \models^*_{\!\relational} }

\newcommand{\sfivemodels}  { \models_{\scriptscriptstyle{\sfive}} }
\newcommand{\notsfivemodels}  { \not\models_{\scriptscriptstyle{\sfive}} }
\newcommand{\KDmodels}  { \models_{\scriptscriptstyle{\KD}} }

\newcommand{\htmodels}  { \ensuremath{ \models_{\scriptscriptstyle{\HT}} } }
\newcommand{\nothtmodels}  { \ensuremath{ \not\models_{\scriptscriptstyle{\HT}} } }

\newtheorem{example}{Example}
\newtheorem{definition}{Definition}
\newtheorem{rem}{Remark} 
\newtheorem{fact}{Fact}
\newtheorem{lem}{Lemma}
\title{Pearce's Characterisation in an Epistemic Domain}
\author{Ezgi Iraz Su
\institute{Sinop University, Department of Computer Engineering, Sinop, Turkey}
\email{eirazsu@sinop.edu.tr}
}
\def\titlerunning{Pearce's Characterisation in an Epistemic Domain}
\def\authorrunning{E.I. Su}
\begin{document}

\maketitle

\begin{abstract}
Answer-set programming ($\ASP$) is a successful problem-solving approach in logic-based $\AI$. 
In $\ASP$, problems are represented as declarative logic programs, and solutions are
identified through their answer sets. 
Equilibrium logic ($\EL$) is a general-purpose nonmonotonic reasoning formalism, 
based on a monotonic logic called here-and-there logic.  
$\EL$ was basically proposed by Pearce as a foundational framework of $\ASP$. 
Epistemic specifications ($\ES$) are extensions of $\ASP$-programs 
with subjective literals. These new modal constructs in the $\ASP$-language make it possible to check whether a regular literal
of $\ASP$ is true in every (or some) answer-set of a program. $\ES$-programs 
are interpreted by world-views, which are essentially 
collections of answer-sets. (Reflexive) autoepistemic logic is a nonmonotonic formalism, 
modeling self-belief (knowledge) of ideally rational agents. 
A relatively new semantics for $\ES$ is based on a combination of 
$\EL$ and (reflexive) autoepistemic logic. 
In this paper, we first propose an overarching framework in the epistemic $\ASP$ 
domain.  
We then establish a correspondence between existing (reflexive) (auto)epistemic 
equilibrium logics and our easily-adaptable comprehensive framework, 
building on Pearce's characterisation of answer-sets as equilibrium models. 
We achieve this by extending Ferraris' work on answer sets 
for propositional theories to the epistemic case and reveal the relationship 
between some $\ES$-semantic proposals.
\end{abstract}

%-------------------------------------------------------------------------------------------------------------%
\section{Introduction}
\label{sec:Introduction}
%-------------------------------------------------------------------------------------------------------------%
\emph{Answer-set programming} ($\ASP$), introduced by Gelfond\&Lifschitz
\cite{GelfondL88,GelfondL91}, is an approach to declarative logic programming. Its reduct-based semantics is defined by \emph{stable models} (alias, \emph{answer-sets}),
essentially the supported classical models of a logic program.
$\ASP$ has demonstrated success
in solving problems within logic-based $\AI$: a problem is first encoded
as a logic program, and then efficient $\ASP$-solvers are
employed to compute its stable models corresponding to the solutions.
However, as Gelfond pointed out in his seminal work \cite{Gelfond91strong},  
$\ASP$ encounters challenges in accurately representing and 
reasoning about incomplete information. The difficulty
arises when a program involves multiple stable models, and
a proposition holds in one stable model but contradicts another. 
The main reason for this drawback lies in the local performance of $\ASP$'s negation as failure ($\texttt{NAF}$) operator, 
which handles incomplete information within individual stable models.
To address this issue and
enable collective reasoning about incomplete information,
we need additional tools in the language of $\ASP$.
Epistemic modal operators provide one potential solution to $\ASP$'s limitation 
with incomplete information. By integrating such operators
into the $\ASP$-language, the new modal constructs in the extended language allow us
to quantify over a collection of stable models and check whether a proposition
holds in every (some) stable model.

The initial approach to this problem is by Gelfond's 
\emph{epistemic specifications}
\cite{Gelfond91strong,gelfond94}, referred to as $\ESdoksandort$ here: 
Gelfond extended $\ASP$ with epistemic constructs 
known as \emph{subjective literals}. Indeed, 
with the incorporation of epistemic modalities $\K$ and $\M$,  
he could represent incomplete information within stable-model collections. 
While a subjective literal $\K\,l$ ($\M\,l$) makes it possible to check 
whether a literal $l$ is true 
in every (some) stable model of a collection, in particular, 
the epistemic negation $\lpnot \K$ accurately captures collective reasoning of incomplete information.
The extended language is interpreted in terms of
\emph{world-views}, which are, in essence, stable-model collections.
However, researchers have soon realised that $\ESdoksandort$ allows unsupported world-views.
Thus, Gelfond himself \cite{gelfond2011new}, along with many other researchers,
have proposed various semantic revisions for $\ES$; each aiming to eliminate
newly-appearing unintended results.
The first counter-example that undermines the soundness
of $\ESdoksandort$-semantics is the model $\set{\set p}$ resulting from 
the epistemic rule $p \leftarrow \K p$. This problem with recursion through $\K$ 
arises due to epistemic circular justification; 
yet efforts to resolve this problem do not focus on the core 
reasons for the emergence of unsupported models in $\ESdoksandort$.
This situation leads to incrementally more complex reduct definitions. Although 
we refrain from calling these solutions ad hoc, as they can be based on reasonable grounds,
we find it crucial to reveal the underlying reasons behind
the existence of such models under $\ESdoksandort$-semantics. Moreover, we introduce a conventional and straightforward generalisation of ASP’s 
reduct definition to epistemic logic programs,
which constitutes our first contribution here.

One line of world-view computing methods in the literature 
depends on the reduct-based 
fixed-point techniques within the logic programming domain, 
with $\ESdoksandort$ serving
as the prototype and most subsequent formalisms being its follow-ups. 
In a parallel, purely logical context, world-views are computed as (reflexive) (auto)epistemic 
extensions of equilibrium models. The initial attempt in this direction 
was made by Wang\&Zhang \cite{WangZ05nested}, whose semantics 
has captured the world-views of $\ESdoksandort$. 
Sequentially, stronger formalisms followed \cite{SuAI20,Cabalar20,Su2021}.
These epistemic equilibrium logics ($\EEL$s) share a common approach: a twofold
world-view computation process. First, they determine stable models of an 
$\ES$-program $\epispec$ in terms of truth ($\texttt{t}$) by applying the 
$\texttt{t}$-minimality criterion of the formalism.
This involves generalising the usual $\texttt{t}$-minimality method which is used to
compute stable models (equilibrium models) 
to $\ES$-programs, resulting in the epistemic equilibrium models ($\EEMs$) 
of $\epispec$.
The inclusion of epistemic constructs into the $\ASP$-language requires the minimisation of these concepts as well, which is fundamental in nonmonotonic epistemic logics. Thus, once $\texttt{t}$-stable models are determined, a 
knowledge-minimality technique should also be applied to guarantee stability
in terms of knowledge ($\texttt{k}$). As a result, world-views are stable-models w.r.t.\ 
both truth and knowledge.
One formally strong $\texttt{k}$-minimality approach applied to $\ES$ is Schwarz's
\cite{Schwarz92} minimal model reasoning for nonmonotonic modal logics. Cabalar et al.\ \cite{Cabalar20} pioneered the introduction of this technique to $\ES$,
proposing a new semantics based on a combination of Pearce's 
equilibrium logic ($\EL$) \cite{Pearce06}
and Schwarz's nonmonotonic $\KD$ \cite{Schwarz92,SU.FI20} 
(equivalently, Moore's autoepistemic logic). Their
formalism so represents a nonmonotonic epistemic logic of belief where
$\K$ is interpreted as \emph{the self-belief of a rational agent}. 
It also captures $\ESdoksandort$-semantics under a foundedness restriction.
Su \cite{Su2021} then suggested employing 
the reflexive closure of $\KD$-models, namely $\SW$-models, 
in the search for $\texttt{k}$-minimal models and proposed reflexive autoepistemic $\EL$. This formalism alternatively applies Schwarz's minimal model technique for nonmonotonic $\SW$ as a $\texttt{k}$-minimality criterion, 
aligning it more closely with other $\ES$-formalisms where
the $\K$ operator formalises \emph{knowledge}. 

The existence of many $\ES$-formalisms without a common agreement makes it difficult
to understand the current state of the art. Thus, as a natural continuation, we explore the relationship between them.
Our reference point will be classifying $\ES$-formalisms 
under a twofold world-view computation method. We then generalise Ferraris' lemma, enabling
\emph{the capture of equilibrium models of a theory as its stable models}, to the epistemic case. Using our new result, we transform $\EEMs$ to truth-stable ($\texttt{t}$-stable) models of epistemic $\ASP$ and vice versa. This work will then help $\ASP$ programmers
better understand existing $\EEL$s, being reflected in the 
logic programming context and also give rise to a versatile and solid framework
in epistemic $\ASP$, an approach not studied before, which will be 
our main contribution here. 
 
The rest of this paper is organised as follows:
Sect.\,\ref{sec:background ASP and ES} 
provides preliminary information about $\ASP$ and Gelfond's primary
$\ESdoksandort$-semantics.
Sect.\,\ref{sec:EpisASP} presents epistemic $\ASP$ ($\EASP$) as a unifying 
framework for several $\ES$-semantics.
Sect.\,\ref{sec:epistemic equilibrium logic} makes a short overview of the existing 
$\EEL$s in the literature, focusing on their $\texttt{t}$-minimality methods.
Sect.\,\ref{sec:correspondence} establishes a correspondence
between these $\EEL$s and $\EASP$ by generalising Ferraris' lemma to $\EASP$.
Sect.\,\ref{sec:conclusion} concludes the paper with
future work plan.

%------------------------------------------------------------------------------------------------------------%
\section{Background: \texorpdfstring{$\ASP$}{ASP} 
and epistemic specifications (\texorpdfstring{$\ES$}{ES}) in a nutshell}
\label{sec:background ASP and ES}
%------------------------------------------------------------------------------------------------------------%
In this context, $\ASP$-formulas are built from an infinite set $\Prop$ of atoms 
using the connectives, viz.\ reversed implication ($\leftarrow$), disjunction ($\lor$), 
conjunction ($\land$), $\naf$ ($\notlp$), strong negation ($\classnot$), true ($\top$) and false ($\bot$). In $\ASP$,
a \emph{literal} $l$ is an atom $p$ or a strongly-negated atom $\classnot p$ 
for $p \in \Prop$. An $\ASP$-program consists of a finite set of rules $\mathtt{r} : \mathtt{head(r)} \leftarrow \mathtt{body(r)}$ s.t.\ $\mathtt{body(r)}$ is formed by a conjunction of literals possibly preceded by $\naf$, 
and $\mathtt{head(r)}$ is formed
by a disjunction of literals: for \, $0 \leq m \leq n \leq k$,
\begin{align}
\label{defn:ASP rules}
l_1 \lor  \ldots \lor l_m ~\leftarrow~ 
l_{m+1} \,\land \, \ldots \, \land \,  l_n \, \land \, \lpnot l_{n+1} \, 
\land \, \ldots \, \land \,\lpnot l_k \,. 
\end{align}%
Alternatively, we call $\Body$ \emph{goal} and its conjuncts \emph{subgoals}.
When $m=0$, we suppose $\mathtt{head(r)}$ to be $\bot$ and call the rule $\mathtt{r}$
\emph{constraint}. When $k=m$, we call $\mathtt{r}$ \emph{fact} and omit both
$\mathtt{body(r)}$ and $\leftarrow$. When $k=n$, we call $\mathtt{r}$ a positive rule. 
A program composed of only positive rules is positive. 
Finally, as strong negation can be removed from a logic program via auxiliary atoms, this paper mostly ignores $\classnot$ for simplicity.

A \emph{valuation} is a consistent (possibly empty) 
set $T$ of literals, i.e., $p \not\in T$ or $\classnot p \not\in T$ for any $p \in \Prop$. 
A valuation $T$ satisfying an $\ASP$-program $\epispec$ (which means $T \models \epispec$) is a \emph{classical} model of $\epispec$. 
Then, 
stable-models of $\epispec$ are 
its reduct-based minimal classical models.
Stable-model semantics is based on a program transformation
that aims to eliminate \onequote{$\notlp$}
from $\epispec$ w.r.t.\ $\epispec$'s 
classical model $T$ (a candidate model), 
resulting in a positive program $\epispec^T$ referred to as 
\emph{reduct} of $\epispec$ w.r.t.\ $T$:
(\emph{reduct-taking}) replace 
$\notlp p \text{ with } \top$ if $T \models \lpnot p$ 
(equivalently, if $T \not\models p$, i.e., $p \not\in T$);
otherwise, with $\bot$.
This approach also requires that the valuation $T$ be a smallest (minimal) 
model of this reduct $\epispec^T$ w.r.t.\ subset relation. Eventually,
the successful models of this process are called \emph{stable models}
(alias, \emph{answer-sets}) of $\epispec$.

%-----------------------------------------------------------------------------------------------------------%
\subsection{Gelfond's epistemic specifications: 
\texorpdfstring{$\ESdoksandort$}{ESdoksandort}}
\label{sec:ES94}
%------------------------------------------------------------------------------------------------------------%
Epistemic specifications ($\ES$) extends $\ASP$-programs with
the epistemic modal operators 
$\K$ (\onequote{known}) and $\M$ (\onequote{may be true}).
The language $\lang{\scriptscriptstyle{\ES}}$ contains four kinds of literals:
\emph{objective literals ($l$)},
\emph{extended objective literals ($L$)},
\emph{subjective literals ($g$)}, and
\emph{extended subjective literals ($G$)}, viz.\ for $p \in \Prop$,
$$
\begin{array}{c c c c}
\pmb{l}   ~~~~ &
\pmb{L}  &
~~~~~~\pmb{g} &
\pmb{G}
\\ \hline 
 p \mid \classnot p ~~   & 
~~~~~~ l \mid \lpnot l       & 
~~~~~~ \K\, l \mid \M\, l  & 
~~~~~~ g \mid \lpnot g
\end{array} 
$$%
Note that $\ASP$'s regular literals are called objective literals in $\ES$. 
By convention, the belief operator $\M$ can be defined in terms of the knowledge operator $\K$, i.e., 
$\M \eqdef \lpnot \K \, \lpnot$, meaning that they are dual.

An \emph{$\ES$-rule} \ \onequote{$\mathtt{r} : l_1 \lor  \ldots \lor  l_m \leftarrow e_{1} ~\land~ \ldots \, \land \, e_n$} \  
is an extension of an $\ASP$-rule (\ref{defn:ASP rules}) with extended subjective literals 
that can appear exclusively in $\mathtt{body(r)}$ as subgoals. Thus,
$\mathtt{body(r)} = e_1 \land \ldots \land e_n$ is a conjunction of arbitrary 
$\ES$-literals. 
Then, an \emph{$\ES$-program} is a finite collection of $\ES$-rules.

%-------------------------------------------------------------------------------------------------------%
\paragraph{Truth conditions:} 
\label{sec:ES94 Truth conditions}
%--------------------------------------------------------------------------------------------------------%
Let $\mathcal{T}$ be a non-empty collection of valuations.
Let $I$ be a valuation, which is not necessarily included in $\mathcal T$. Then, 
for an objective literal $l$ and a subjective literal $g$, we have:
$$\begin{array}{lcllcl}
\mathcal T, I \models l & \text{ if }& l\in I; \hspace{104pt}
\mathcal T, I \models \lpnot l & \text{ if }&\!\! l \notin I .
 \smallskip
\\
\mathcal{T},I \models \K\,l & \text{ if }& l \in T \text{ for every } T \in \mathcal{T};
\hspace{30pt}
\mathcal{T},I \models \lpnot g & \text{ if }& \mathcal{T}, I \not\models g.
\\
\mathcal{T}, I \models \M\,l & \text{ if }& l \in T \text{ for some } T \in \mathcal T;
&&
\end{array}$$

Note that the satisfaction of an objective literal $l$ is independent of $\mathcal T$,
while the satisfaction of a subjective literal $g$ is independent of $I$. 
Thus, we simply write $\mathcal T \models g$ or $I \models l$.
Then, we define the satisfaction of an $\ES$-program $\epispec$ as follows:
$\mathcal T, I \models \epispec$ \ if \ for every rule $\mathtt{r} \in \epispec$, \,
$\mathcal T, I \models \mathtt{r}$, i.e., explicitly 
\begin{align*}
\mathcal T, I \models \mathtt{body(r)} \text{~~ implies~~ }
\mathcal T, I \models \mathtt{head(r)}.
\end{align*}

An $\sfive$-model is a nonempty collection of possible worlds, each with assigned truth values, where the connection between these worlds is defined by an equivalence relation (reflexive, symmetric, and transitive).
In this context, we assume an $\sfive$-model $\mathcal T$ 
to be in the form of a nonempty set of valuations s.t.\ 
any two valuations are related.
When $\mathcal T, T \models \epispec$
for every $T \in \mathcal T$,
we say that $\mathcal T$ is a \emph{classical} $\sfive$-\emph{model} 
of $\epispec$. 
In particular, when we designate a valuation $T$ s.t.\ $\mathcal T, T \models \epispec$, we call $(\mathcal T, T)$ a \emph{pointed} $\sfive$-model
of $\epispec$. Extending this to a set $\mathcal T_0$ of designated valuations,
$\tuple{\mathcal T, \mathcal T_0}$ is said to be a \emph{multi-pointed} $\sfive$-model of $\epispec$. 
To facilitate reading, we symbolise a multi-pointed 
$\sfive$-model $\tuple{\mathcal T, \mathcal T_0}$ 
by underlying its designated valuations $T \in \mathcal T_0$ 
in an explicit representation of $\mathcal T$.
Given $\mathcal T= \set{\set a, \set b , \emptyset}$, the (multi)pointed $\sfive$-models
$\tuple{\mathcal T, \set a}$ and $\tuple{\mathcal T, \set{\set a,\set b}}$ correspond to
$\set{\underline{\set a}, \set b, \emptyset}$ and
$\set{\underline{\set a}, \underline{\set b}, \emptyset}$ respectively. 
When no valuation is underlined or specified, by default this means that any valuation of $\mathcal T$ behaves as designated. 
The rest of the paper uses the terms \doublequotes{point}, \doublequotes{valuation} and 
\doublequotes{world} interchangeably. Finally,
given a syntactic $\ES$-construct (head, rule, program, etc.) 
$\phi$, when $\mathcal T, T \models \phi$ for every $T \in \mathcal T$, we simply
write $\mathcal T \models \phi$. 

%-----------------------------------------------------------------------------------------------------%
\paragraph{Semantics:} 
\label{sec:ES94 Semantics}
%-----------------------------------------------------------------------------------------------------%
An $\ES$-program $\epispec$ is interpreted by means of its world-views, which are 
selected from among its $\sfive$-models. Thus,
given a candidate $\sfive$-model $\mathcal T$ of $\epispec$,
we first compute the (epistemic) reduct $\epispec^\mathcal{T} = \{ \mathtt{r}^\mathcal{T} : 
\mathtt{r} \in \epispec \}$
of $\epispec$ w.r.t.\ $\mathcal{T}$ by
replacing every subjective literal $\K\,l$ ($\M\,l$), possibly preceded by $\naf$, 
with $\top$ if  $\mathcal T \models \K\,l \, (\M\,l)$; otherwise, with $\bot$.
Then, $\mathcal{T}$ is a \emph{world view} of $\epispec$ if
$\mathcal T = \anset(\epispec^\mathcal T)$ where
$\anset(\epispec)$ denotes the set of all stable models of $\epispec$.
The reduct definition of $\ESdoksandort$
is so oriented to remove extended subjective literals.
The resulting program $\epispec^{\mathcal T}$
is then a nonepistemic, but not necessarily positive
$\ASP$-program potentially containing $\naf$. In fact, $\ESdoksandort$
offers a twofold reduct definition; first removing epistemic operators 
w.rt.\ $\mathcal T$ and then eliminating $\naf$ w.r.t.\ $T \in \mathcal T \cup \mathcal X$ akin to $\ASP$'s methodology.

%------------------------------------------------------------------------------------------------------------%
\subsubsection{Motivation}
\label{sec:EpisASP}
%------------------------------------------------------------------------------------------------------------%
\begin{example}
\label{ex: motovating ex 1 for EASP} \normalfont
The one-rule program $\Gamma = \set{a \leftarrow \K\,a}$ has 2 world-views,
$\set{\emptyset}$ and $\set{\set a}$ in $\ESdoksandort$. Among these, 
only the former is intended. The self-supported model $\set{\set a}$
appears due to the fact that $\ESdoksandort$-reduct attacks positive 
(not preceded by $\naf$) literals. This approach causes unsupported models to provide
fake derivations for head-literals, which in return produce these models by fixed-point
justifications. Thus, Gelfond's methodology includes flaws 
for programs containing cyclic dependencies like $\Gamma$,
$\set{a \leftarrow \K \,a \land \lpnot \K \, b}$, 
$\set{a \leftarrow \K \,a \land \lpnot b}$, etc. 
Such circular scenarios may arise 
when the goal contains a positive subjective literal and 
is satisfied by the candidate unsupported $\sfive$-model. Notice that 
transformation of a literal into true/false w.r.t.\ its truth-value is secure
when it is preceded by $\naf$ with literal reading \emph{there is no evidence}, 
or when there exits logical derivations of literals as used by splitting property 
of (epistemic) $\ASP$. To overcome this problem, Gelfond \cite{gelfond2011new}
slightly modifies his reduct definition by replacing $\K\,p$ with $p$ 
when $\mathcal T \models \K\, p$
and partly avoids circular justifications,
but the problem of recursion via $\M$ prevails.
\end{example}
\begin{table}
\centering
\small{
\caption{Kahl's reduct definition proposed in his PhD thesis \cite{kahl2014refining} 
with changes over \cite{gelfond2011new} in bold.} 
\label{table:Kahl.reduct}
\begin{tabular}{l l l l l l l}
\hline
            literal $G$ 	~~
        & if  $\mathcal T \models G$ 
        & if  $\mathcal T \not\models G$ &&
                    literal $G$ 	~~
        & if  $\mathcal T \models G$  
        & if  $\mathcal T \not\models G$
\\ \hline
$\K\,l$ 		&  replace by $l$	   & replace by $\bot$  &&
$\lpnot \K\,l$ 			& replace by $\top$		& \textbf{replace by} $\boldsymbol{\lpnot l}$	
\\ \hline
$\M\,l$ 			& replace by $\top$	   &  \textbf{replace by} $\boldsymbol{\lpnot \lpnot l}$ &&
$\lpnot \M\,l$ 	&  \textbf{replace by} $\boldsymbol{\lpnot l}$  & replace by $\bot$
\\ \hline
\end{tabular} }
\end{table}

This modification has probably necessitated further changes in his reduct definition 
as shown in Table\ref{table:Kahl.reduct}. The underlying reasons of Kahl's new reduct 
\cite{kahl2014refining} may be grounded as follows: 
(1) If $\mathcal T \not\models \lpnot\K\, l $, then $\mathcal T \models \K\, l$. When
the reduct definition transforms $\K\,l$ into $l$, it  
replaces $\lpnot \K\, l$ with $\lpnot l$. 
(2) Remember that $\M \eqdef \lpnot \K\,\lpnot$. If $\mathcal T \not\models \M\, l$, then
$\mathcal T \not\models \lpnot \K\, \lpnot l$, i.e., $\mathcal T \models \K\, \lpnot l$. 
A similar reasoning may force the transformation of $\K\,\lpnot l$ into $\lpnot l$; $\M\, l$ into $\lpnot \lpnot l$. 
(3) If $\mathcal T \models \lpnot \M\,l$, then $\mathcal T \models \K\,\lpnot l$. 
If $\K\,\lpnot l$ is transformed into  $\lpnot l$, then $\lpnot \M\,l$ is turned into
$\lpnot \lpnot \lpnot l$, equivalently \cite{ASP-strong.equivalence} 
into $\lpnot l$. While this explanation is a guess,
in fact when $\naf$ is involved, such further intricate changes may not be required.
\begin{example}
\label{ex: motovating ex 2 for EASP} \normalfont
Another recursive program $\Sigma =\set{a \leftarrow \M\,a}$ yields the same world-views in $\ESdoksandort$. Researchers have widely varying perspectives on the intended models of $\Sigma$.
While some find both models reasonable, the others argue that $\Sigma$ should 
have one model; yet they also differ on which model should be preferred. We will not engage in
this debate, as different approaches may prove useful depending on the specific problem 
at hand. Our stance on the topic is distinct. In alignment with Su et al.'s
approach \cite{SuAI20}, and following the tradition of intuitionistic modal logics, 
we will adopt a positive belief operator here, namely $\Khat$, 
which is not definable in terms of $\K$ and $\notlp$. As $\M \eqdef \lpnot \K\,\lpnot$, in our opinion,
$\M$ cannot be regarded as purely positive like $\lpnot\lpnot a$ in $\ASP$. Remember that Su et al.\ handle $\M$ as a syntactic sugar, giving a concise representation for the equivalent formulas 
$\lpnot \lpnot \Khat$, $\Khat\,\lpnot\notlp$, and $\lpnot \K\,\lpnot$. 
Also recall that in epistemic $\ASP$, aligning with  
$\ASP$, double $\naf$
should not vanish regardless of where it occurs. On the other hand, 
similarly to $\Gamma$ in Ex.\,\ref{ex: motovating ex 1 for EASP},
we claim that the intended model of $\Sigma'=\set{a\leftarrow \Khat\, a}$ should
be $\set{\emptyset}$.
\end{example}

%-----------------------------------------------------------------------------------------------------------%
\section{Epistemic Answer Set Programming (\texorpdfstring{$\EASP$}{EASP})}
\label{sec:EpisASP}
%-----------------------------------------------------------------------------------------------------------%
This section introduces a direct generalisation of logic programs under stable-model
semantics (aka, $\ASP$-programs) to epistemic logic programs 
under stable $\sfive$-model semantics. This new concept has been partially explored 
by \cite{Su19jelia}.
The shift from the general term \emph{world-view} to
\emph{stable $\sfive$-model} in $\EASP$, and \emph{equilibrium $\sfive$-model} 
in the following section is intended to emphasise the purpose of this work.
Our main motivation for this study arises from the unsupported models that emerge
due to circular justifications under $\ESdoksandort$-semantics 
(see Ex.\,\ref{ex: motovating ex 1 for EASP}-\ref{ex: motovating ex 2 for EASP}).
$\ESdoksandort$'s reduct definition deviates somewhat from the traditional approach.
We here propose a new reduct definition for $\ES$-programs,
oriented to eliminate exclusively $\naf$. Thus, our reduct is a positive program, 
similar to the method in search for stable models. 

The new approach exploits a two-step computation process, 
focusing on stability in terms of truth ($\texttt{t}$) and knowledge ($\texttt{k}$). 
The method involves finding the minimal models in terms of truth first, and then 
refining them further w.r.t.\ a $\texttt{k}$-minimality criterion 
to select stable $\sfive$-models. Such models then capture truth and knowledge minimality concepts that is central in (nonmonotonic) epistemic $\ASP$. 
In broader terms, 
what we refer to as $\texttt{t}$-minimality in $\ES$ is essentially an extension
of the familiar minimisation criterion of $\ASP$ from classical models to classical 
$\sfive$-models. However, $\texttt{k}$-minimality is a relatively new concern 
within the $\ASP$ field compared to the well-established method of $\texttt{t}$-minimality. 
The necessity for such a technique has become evident 
with the incorporation of epistemic 
concepts into $\ASP$ and the need to 
maximise epistemic possibilities (i.e., ignorance).

A stable $\sfive$-model $\mathcal T$ of an epistemic logic program $\epispec$ 
is its $\sfive$-model 
s.t.\ each valuation $T \in \mathcal T$ forms $\epispec$'s
pointed $\sfive$-model $(\mathcal T, T)$ where $T$ is minimal
w.r.t.\ truth and $\mathcal T$ is minimal w.r.t.\ knowledge.
For a nonepistemic $\ASP$-program $\epispec$, such valuations are $\epispec$'s
stable-models in $\ASP$, and the (unique) stable $\sfive$-model $\mathcal T$ is the set of all such models.
Similar to stable-models of $\ASP$, the intuition underlying stable $\sfive$-models  
is to capture the \emph{rationality} 
of an agent associated with an epistemic logic program
$\epispec$:
\doublequotes{\emph{an agent is not supposed to believe 
anything that it is not forced to believe.}}
The aim, in principle, is to determine which propositions can be 
nonmonotonically
inferred from $\epispec$ by considering all its stable-models.
These inferences are then used to deduce new information 
about the knowledge of $\epispec$. 

%--------------------------------------------------------------------------------------------------%
\subsection{The Language of \texorpdfstring{$\EASP$}{EASP}
 ($\lang{\scriptscriptstyle{\EASP}}$) }
\label{sec:Lang.EpisASP}
%---------------------------------------------------------------------------------------------------%
The language $\lang{\scriptscriptstyle{\EASP}}$
extends that of $\ASP$ by epistemic modalities $\K$ and $\Khat$.
Literals ($\lambda$) of $\lang{\scriptscriptstyle{\EASP}}$ are of two types;
\emph{objective ($l$)} and
\emph{subjective ($g$)} literals, viz.\,
$l \bnf p ~\mid~ \classnot p$ \, and \,
$g \bnf  \K\,l~\mid~ \Khat\,l$
for $p \in \Prop$. Then, $\lpnot \lambda$
means \emph{failing to derive $\lambda$, the query $\lambda ?$ is undetermined and 
assumed to be false}; yet we do not offer literal interpretations
of the modalities for the sake of flexibility.

Replacing literals of $\ASP$ with those of $\EASP$ in (\ref{defn:ASP rules}), we obtain
an \emph{$\EASP$-rule} $\mathtt{r}$, viz.\ 
\begin{align} \label{EASP rules}
\lambda_1 \lor \ldots \lor \lambda_m \,\leftarrow\, 
\lambda_{m+1} ~\land~ \ldots ~\land~ \lambda_n ~\land~
\lpnot \lambda_{n+1} ~\land~ \ldots ~\land~ \lpnot \lambda_k\,. ~~~~~~
(\text{ for } 0 \leq m \leq n \leq k)
\end{align}
in which $\lambda_i$'s are objective or subjective literals for every $i=1,\ldots,k$.
When we restrict $\lambda_i$'s to objective literals, the resulting program 
is a disjunctive logic program \cite{GelfondL91}.
Hence, $\EASP$-rules are conservative extensions of $\ASP$'s disjunctive rules (\ref{defn:ASP rules}).
Different from $\ES$, we allow $\K\,l$ and $\Khat\,l$ to appear in $\mathtt{head(r)}$.
While the use of subjective literals in the head has not yet been fully explored, we still 
find it useful to provide the same syntax structure with $\ASP$
for easier understanding of the approach. This way, extensions to richer
languages are straightforward via the main $\ASP$ track.
An \emph{epistemic logic program} ($\ELP$), also known as $\EASP$-program, is a finite collection of $\EASP$-rules (\ref{EASP rules}). 

%--------------------------------------------------------------------------------------------------------%
\subsection{Semantics of \texorpdfstring{$\EASP$}{EASP} in terms of stable 
\texorpdfstring{$\sfive$}{sfive}-models}
\label{sec:Semantics.EpisASP.epistemic stable models}
%--------------------------------------------------------------------------------------------------------%
We first introduce $\texttt{t}$-minimality concept in $\EASP$. Based on the existing 
$\ES$-formalisms in the literature, we provide two slightly different approach.
For example, the program $\Phi=\set{\mathtt{r}_1,\mathtt{r}_2,\mathtt{r}_3}$
\begin{align}
\label{ex: t-minimality slight difference}
\mathtt{r}_1 = a \lor b. \hspace{5em}
\mathtt{r}_2 = a \leftarrow \K\,b. \hspace{5em}
\mathtt{r}_3 = b \leftarrow \K\,a.
\end{align}
may produce $\texttt{t}$-minimal models 
$\mathcal T_1 = \set{ \set a , \set b }$ and $\mathcal T_2 = \set{ \set{a, b} }$; yet
it may also yield $\mathcal T_1$ only, depending on how restrictive we want to be.
In $\EASP$, this subtle distinction originates from differing approaches
of $\texttt{t}$-minimality techniques, 
emphasising functional vs.\ relational perspective.
\begin{definition}[\textbf{weakening of a point in an $\sfive$-model
in terms of truth: functional approach}]
\label{defn:weakening w.r.t. truth-functional} \normalfont
Given a nonempty collection $\mathcal T$ 
of valuations, let $\weak {\suchthat} \mathcal T \rightarrow 2^{\Prop}$ 
be a \emph{subset} 
function s.t.\ $\weak(T) \subseteq T$ for every $T \in \mathcal T$. 
Let $id$ refer to the identity function, and let
$\weak[\mathcal T]=\bigset{\weak(T)}_{\scriptscriptstyle{T\in \mathcal T}}$
denote the image of $\mathcal T$ under $\weak$.
A \emph{functional (\functional) weakening} of $\mathcal T$ 
at a point $T \in \mathcal T$ 
by means of $\weak$ is identified with $\tuple{\weak[\mathcal T], \weak(T)}$
s.t.\ $\weak \neq id$ on $\mathcal T$ and
$\restr{\weak}{ \mathcal T \setminus \set T}= id$, by which we take a strict subset of 
$T \in \mathcal T$ and keep the elements of $\mathcal T \setminus \set{T}$ unchanged.
We say that $\tuple{\weak[\mathcal T], \weak(T)}$ is \emph{\functional-weaker} 
than $\tuple{\mathcal T, T}$
on $T \in \mathcal T$ and denote it by $\tuple{\weak[\mathcal T], \weak(T)} \lhd_{\functional} \tuple{\mathcal T, T}$. 
\end{definition}

Def.\,\ref{defn:weakening w.r.t. truth-functional} has already been introduced by \cite{Su19jelia}; yet the following more cautious approach is novel.
\begin{definition}[\textbf{weakening of a point in an $\sfive$-model
in terms of truth: relational approach}]
\label{defn:weakening w.r.t. truth-relational} \normalfont
Let $\weakrelational \suchthat \mathcal T \Rightarrow 2^{\Prop}$ 
be a multi-valued \emph{subset} function s.t.\ 
$\weakrelational(T) \subseteq 2^T$ and
$\weakrelational(T) \neq \emptyset$ for every $T \in \mathcal T$. 
For ease of understanding,
we also design $\weakrelational$ as a serial subset relation, 
relating each $T \in \mathcal T$ to at least one element from $2^T$ 
and form the collection
$\weakrelational=\bigset{(T,H) :H \in \weakrelational(T)}_%
{\scriptscriptstyle{T\in \mathcal T}}$. 
Then, a \emph{relational (\relational) weakening} of $\mathcal T$ 
at a point $T \in \mathcal T$ by means of $\weakrelational$ is identified
with $\tuple{\weakrelational[\mathcal T], \weakrelational(T)}$
s.t.\ $\weak \neq id$ on $\mathcal T$ and
$\restr{\weak}{ \mathcal T \setminus \set T}= id$, by which we replace only
$T$ in $\mathcal T$ by a set of its subsets including 
at least one strict subset $H \subset T$.
We say that $\tuple{\weakrelational[\mathcal T], \weakrelational(T)}$ 
is \emph{\relational-weaker} than $\tuple{\mathcal T, T}$
on $T \in \mathcal T$ and denote it by $\tuple{\weakrelational[\mathcal T], \weakrelational(T)} \lhd_{\relational} \tuple{\mathcal T, T}$.
\end{definition}

We now define a nonmonotonic satisfaction relation 
$\starmodels$ for $\sfive$-models, involving
a $\texttt{t}$-minimality criterion based on set inclusion
over each set $T \in \mathcal T$.
Intuitively, this condition says that
none of the weakenings of $\tuple{\mathcal T,T}$ is an $\sfive$-model
of an epistemic logic program ($\ELP$) $\epispec$ for every $T \in \mathcal T$.
\begin{definition} [\textbf{generalisation of the truth-minimality ($\texttt{t}$-minimality) criterion of $\ASP$ to $\EASP$}]
\normalfont
For a positive $\EASP$-program $\epispec$,
let $\mathcal T$ be a nonempty collection of valuations, and $T \in \mathcal T$.
Then, we have:
\begin{align}
\label{defn:truth minimality}
\mathcal T, T \starmodels_{\!\functional} \epispec \text{ \ \ iff \ \ }
\mathcal T, T \models \epispec \text{ ~and~ }
\weak[\mathcal T], \weak(T) \not\models \epispec
\text{ for every }  \weak  \text{ s.t.\ } 
\tuple{\weak[\mathcal T], \weak(T)} \lhd_{\functional} \tuple{\mathcal T, T}.
\end{align}
Thus, $\mathcal T$ is a \emph{$\tfunctional$-minimal} model of $\epispec$ if
$\mathcal T, T \starmodels_{\!\functional} \epispec$ for every $T \in \mathcal T$
\cite{Su19jelia}. In this paper, we also define
$\starmodels_{\!\relational}$ by replacing $\weak$ with $\weakrelational$, and
$\lhd_{\functional}$ with $\lhd_{\relational}$ 
in (\ref{defn:truth minimality}) and produce 
\emph{$\trelational$-minimal} models of $\epispec$ accordingly.
\end{definition}

Although the above definitions seem to be technically complex and daunting, 
they are easily applied:
\begin{example}
\label{ex: t-minimality slight difference cont'd} \normalfont
Reconsider first the program $\Phi$, 
identified by (\ref{ex: t-minimality slight difference}), 
and its $\sfive$-model $\mathcal T_2=\set{\set{a,b}}$. Then 
construct $2^{\set{a,b}} = \set{\set{a,b}, \set a, \set b, \emptyset}$. Since the
$\functional$-weaker models $\set{\set a}$, $\set{\set b}$, and $\set{\emptyset}$ of $\mathcal T_2$
do not satisfy $\mathtt{r_3}$, $\mathtt{r_2}$, and $\mathtt{r_1}$ respectively, $\Phi$ does not hold in them either. Thus, $\mathcal T_2$ is
a $\tfunctional$-minimal model of $\Phi$.

What eliminates $\mathcal T_2$ in the second approach is the relational nature 
of the weakening methodology because now we have to consider all possible subsets 
of $2^{\set{a,b}}$ different from $\emptyset$ and $\mathcal T_2$, i.e., all the elements of the set $2^{2^{\set{a,b}}}\setminus \set{\mathcal T_2, \emptyset}$. 
The element $\set{\set a , \set b}$
from this set, namely an $\relational$-weakening of $\mathcal T_2$ at the point
$\set{a,b} \in \mathcal T_2$, satisfies $\Phi$. Thus, $\mathcal T_2$ fails to be a $\trelational$-minimal model of $\Phi$.

Note that when we consider $\mathcal T_1$, different from the singleton model
$\mathcal T_2$, we follow the above steps
for every pointed $\sfive$-model of $\mathcal T_1$, viz.\ 
$\set{\underline{\set a}, \set b}$ and $\set{\set a, \underline{\set b}}$. 
Also note that $\Phi$
is a positive program, and its reduct trivially equals itself. 
Thus, our reduct is not interested in
the positive literals $\K\,a$ and $\K\,b$ in $\Phi$.
\end{example}
\begin{fact} \normalfont
Functional minimality implies relational minimality 
because any function can be defined as a relation. 
Thus, a $\trelational$-minimal model of an $\ELP$
$\epispec$ is a $\tfunctional$-minimal model of $\epispec$, 
but not vice versa. 
\end{fact}
\begin{example} 
\label{ex: to understand t-min better}\normalfont
Consider the $\EASP$-program 
$\Sigma=\set{\mathtt{r_1}, \mathtt{r_2}, \mathtt{r_3}, \mathtt{r_4}}$
with its rules explicitly represented below:
\begin{align*}
\mathtt{r_1} = a \lor b.             \hspace{30pt} 
\mathtt{r_2} = c \leftarrow b.   \hspace{30pt} 
\mathtt{r_3} = d \leftarrow \K\,a. \hspace{30pt} 
\mathtt{r_4} = \bot \leftarrow \Khat\,d.
\end{align*}
Note that $\Sigma$ is a positive program.
We compute that $\bigset{\set a, \set {b, c}}$ is a 
$\texttt{t}$-minimal model of $\Sigma$:
$\bigset{\underline{\set a}, \set {b,c}} \models \Sigma$
while its only $\functional$-weakening 
$\bigset{\underline{\emptyset}, \set {b,c}}$ refutes it. Likewise,
$\bigset{\set a, \underline{\set {b,c}}} \models \Sigma$
while all its $\functional$-weakenings, i.e., $\bigset{\set a, \underline{\set {b}}}$, 
$\bigset{\set a, \underline{\set {c}}}$,
and $\bigset{\set a, \underline{\emptyset}}$ do not satisfy it.
We leave it to the reader to show that
$\mathcal T$ is also $\trelational$-minimal; yet
we give a hint that while computing the $\relational$-weakenings of, for example, 
$\bigset{\set a, \underline{\set {b,c}}}$, we consider 
all possible models
including $\bigset{\set a, \underline{\set b},\underline{\set c}}$, $\bigset{\set a, \underline{\set {b,c}}, \underline{\emptyset}, \underline{\set b}}$, etc. There are 14 of such models. 
Clearly, $\set{\set {b,c}}$ is $\Sigma$'s other $\texttt{t}$-minimal model,
that is unintended and to be eliminated under $\texttt{k}$-minimality conditions. 
Note that like $\K\,a$, the other positive literal $\Khat\,d$ is not 
involved in the reduct-taking process. 
\end{example}
\begin{rem} \normalfont
The need for relational minimality arises from the fact that under singleton $\sfive$-models like $\set{\set{p}}$, the literals $\K\,p$, $\Khat\,p$, and $p$ are of no difference since quantification is trivially performed over just one valuation $\set p$.
For instance, notice that when we replace $\K\,l$ by $l$ in $\Phi$ 
(\ref{ex: t-minimality slight difference}), 
the resulting $\ASP$-program has the stable model $\set{a,b}$. 
Using relational weakening, we increase epistemic possibilities 
(points) while reducing truth. Quantifying over these points then reveals the nontrivial functionality of subjective literals.
In a sense, the relational $\texttt{t}$-minimality approach simultaneously embeds in itself a kind of $\texttt{k}$-minimality strategy by increasing ignorance with epistemic possibilities.
The difference between two minimality methods strikingly appears for $\Phi$
under the $\sfive$-model $\set{\set{a,b}}$ (see Ex.\,\ref{ex: t-minimality slight difference cont'd}). 
Adding the constraint 
$\mathtt{r_c}=\bot \leftarrow \lpnot \K\,a$ into $\Phi$, the new program 
$\Phi'=\Phi \cup \set{\mathtt{r_c}}$ has a world-view $\set{\set{a,b}}$ under several $\ES$-formalisms.
Some researchers find this result unsupported; 
yet the existing $\texttt{k}$-minimality techniques is unable to eliminate this model. 
Thus, a more restrictive $\texttt{t}$-minimality tool has been designed
to remove models like $\set{\set{a,b}}$ while computing $\texttt{t}$-minimal models.  
We do not discuss this issue here, as our aim is just to establish a correspondence between existing $\ES$-formalisms; to put it better, to demonstrate the reader how current epistemic equilibrium logics are manifested 
in the logic programming domain.
\end{rem}

We will now see how to compute stable w.r.t.\ truth ($\texttt{t}$-stable) 
models of an arbitrary $\EASP$ program potentially including $\naf$.
Satisfaction of the subjective literal $\Khat\,l$ is the same as $\M\,l$ in $\ES$. 
What makes the
difference is primarily how the reduct definition handles them.
\begin{definition} [\textbf{generalisation of the conventional reduct 
definition of $\ASP$ to $\EASP$}]
\label{defn:epis.reduct} \normalfont  
For an arbitrary $\EASP$-program $\epispec$, let $\mathcal T$ 
be a nonempty collection of 
valuations, and let $T \in \mathcal T$.
Then, the reduct $\epispec^{\tuple{\mathcal T,T}}$ of $\epispec$
w.r.t.\ the pointed $\sfive$-model $\tuple{\mathcal T,T}$ is defined by
replacing every occurrence of $\naf$-negated (i.e., preceded by $\naf$) 
literals $\lpnot \lambda$ in $\epispec$ with the truth-constants
\begin{align*}
\bot \text{ \ if \ } \mathcal T,T &\models \lambda ~~~~~~~
(\text{ for } \lambda=l  \text{ \ if \ } T \models l;~~~ 
\text{ for } \lambda=\K\,l \ (\Khat\,l) \text{  \ if \ } 
\mathcal T \models \K\,l \ (\Khat\, l) \ );
\\
\top \text{ \ if \ } \mathcal T, T &\not\models \lambda  ~~~~~~~
(\text{ for } \lambda=l \text{  \ if \ } T \not\models l;~~~ 
\text{ for } \lambda=\K\,l \ (\Khat\,l) \text{ \ if \ } \mathcal T \not\models \K\,l \ (\Khat\,l) \ ).
\end{align*}
Thus, $\mathcal T$ is a \emph{$\texttt{t}$-minimal model} of 
$\epispec$ \, if \, $\mathcal T, T \starmodels \epispec^{\tuple{\mathcal T, T}}$ 
for every $T \in \mathcal T$ \cite{Su19jelia}.
\end{definition}

While Def.\,\ref{defn:epis.reduct} provides a general definition,
its specialisation to $\tfunctional$ and $\trelational$ is straightforward. When 
these methods do not result in a distinction, we refer to them by
the general name \doublequotes{truth} ($\texttt{t}$).
\begin{example}\label{ex: t-min general example} \normalfont
Consider the $\EASP$-program 
$\Gamma=\set{\mathtt{r_1},\mathtt{r_2},\mathtt{r_3},\mathtt{r_4}}$ where its rules are explicitly shown below:
\begin{align*}
\mathtt{r_1} = a \lor b. \hspace{20pt}
\mathtt{r_2} = c \leftarrow \Khat\,a \land \lpnot b. \hspace{20pt}
\mathtt{r_3} = d \leftarrow \lpnot \K\,a \land b. \hspace{20pt}
\mathtt{r_4} = \bot \leftarrow \lpnot \Khat\,c.
\end{align*}
We claim that $\bigset{\set{a,c},\bigset{b,d}}$ is a $\texttt{t}$-minimal model of $\Gamma$.
We first compute the following reducts:
$$
\left.\begin{matrix}
\hspace{2.6em} a \lor b.   & \\
\hspace{2.65em} c \leftarrow \Khat\,a \land \lpnot \bot. &\\
\hspace{1.65em} d \leftarrow \lpnot\bot \land b. &\\
\bot \leftarrow \lpnot \top.
\end{matrix}\right\} \Gamma^{\set{\underline{\set {a,c}}, \set {b,d}}}
\\\hspace{15pt} \text{ and } \hspace{8pt}
\left.\begin{matrix}
\hspace{2.6em} a \lor b.   & \\
\hspace{2.65em} c \leftarrow \Khat\,a \land \lpnot \top. & \\
\hspace{1.65em} d \leftarrow \lpnot\bot \land b. & \\
\bot \leftarrow \lpnot \top.
\end{matrix}\right\} \Gamma^{\set{\set {a,c}, \underline{\set {b,d}}}}
$$
The above reducts 
are respectively equivalent to
$\set{\mathtt{r_1} \, ,\, c\leftarrow \Khat\,a \, ,\, d\leftarrow b}$ and 
$\set{\mathtt{r_1} \, ,\, d\leftarrow b}$: when $\bot$ ($\lpnot \top$) 
appears as a subgoal, the goal fails to hold. This
means that the effect of the entire rule $\mathtt{r}$ is negligible, 
and $\mathtt{r}$ can be safely omitted. When
$\top$ ($\lpnot \bot$) appears as a subgoal, 
$\top$ can be dropped from the subgoals of $\mathtt{body(r)}$
as it trivially holds.
While
$\bigset{\underline{\set{ a,c}}, \set {b,d}} \models 
\Gamma^{\set{\underline{\set{ a,c}}, \set {b,d}}}$,
all its $\functional$-weakenings, viz.\
$\bigset{\underline{\set{ a}}, \set {b,d}}$,
$\bigset{\underline{\set{ c}}, \set {b,d}}$  and
$\bigset{\underline{\emptyset}, \set {b,d}}$, refute it. While
$\bigset{\set{a,c}, \underline{\set {b,d}}} \models
\Gamma^{\set{\set{a,c}, \underline{\set {b,d}}}}$, all its $\functional$-weakenings, viz.\
$\bigset{\set{ a,c }, \underline{\set {b}}}$,
$\bigset{\set{a,c}, \underline{\set {d}}}$ and
$\bigset{\set{a,c}, \underline{\emptyset}}$, refute it. Finally, notice that
the $\sfive$-model $\set{\set{a,c}}$
is the other (unintended) $\texttt{t}_{\!{\functional}}$-minimal model of $\Gamma$, and
both $\texttt{t}$-minimality tools
produce the identical results for $\Gamma$.
\end{example}

In a parallel, purely logical context, world-views are alternatively 
computed as epistemic extensions of equilibrium models. A first step towards
epistemic equilibrium logic belongs to Wang\&Zhang \cite{WangZ05nested}. 
As their approach has generalised $\ESdoksandort$ and also due to page restrictions, 
we do not include it below.

%--------------------------------------------------------------------------------------------------------%
\section{Epistemic Extensions of Equilibrium Logic}
\label{sec:epistemic equilibrium logic}
%--------------------------------------------------------------------------------------------------------%
\emph{Equilibrium logic} ($\EL$) is a nonmonotonic formalism, basically proposed by Pearce \cite{Pearce06} as a logical and mathematical framework of $\ASP$.
$\EL$ is based on \emph{here-and-there logic} ($\HT$), 
a three-valued monotonic logic which is intermediate between classical logic 
and intuitionistic logic. 
An $\HT$-model is an ordered pair $(H,T)$ of valuations 
$H,T \subseteq \Prop$ satisfying $H \subseteq T$.
The semantics of $\EL$, via \emph{equilibrium models}, is obtained through a
$\texttt{t}$-minimality criterion over $\HT$-models: $T$ is an equilibrium model of
$\phi$ iff $T,T \htmodels \phi$ (i.e., $T \models \phi$)
and (\emph{$\texttt{t}$-minimality condition})
$H,T \nothtmodels \phi$ for any $H$ strictly included in $T$
($H \subset T$). In summary,
Pearce has generalised $\ASP$ by characterising 
its stable-models as equilibrium models in $\EL$. 

%----------------------------------------------------------------------------------------------------------%
\subsection{Su et al.'s approach 
{\normalfont (\texorpdfstring{$\ESyirmia$}{ESyirmia}):} 
autoepistemic equilibrium logic \normalfont (\texorpdfstring{$\AEEL$}{AEEL})}
\label{subsec:EEL.FHS}
%----------------------------------------------------------------------------------------------------------%
Inspired by $\EL$'s success as a foundational framework for $\ASP$, 
Su et al.\ introduced \cite{SuThesis15,Suijcai15,SuAI20} 
an epistemic extension of $\EL$ as an alternative semantics for $\ES$. 
We here name their approach $\ESyirmia$ and recall
how $\ESyirmia$ produces its $\texttt{t}$-minimal models, 
namely epistemic equilibrium models ($\EEMs$). 
For our purposes, we do not include their $\texttt{k}$-minimality method, 
selecting $\ESyirmia$-world-views among its $\EEMs$. 

%-------------------------------------------------------------------------------------------------------%
\subsubsection{Epistemic here-and-there logic {\normalfont (\texorpdfstring{$\EHT$}{EHT})} 
and its equilibrium \texorpdfstring{$\sfive$}{sfive}-models
w.r.t.\ truth}
\label{subsubsection:EHT.FHS}
%--------------------------------------------------------------------------------------------------------%
$\EHT$ extends $\HT$ with nondual epistemic modalities $\K$ and $\Khat$, both of which are primitive and structurally identical to the modalities in $\EASP$. 
Depending on knowledge-minimality conditions, 
these modalities may characterise different epistemic concepts, so
we do not assign them a literal reading for generality.
The language of $\EHT$ ($\lang{\scriptscriptstyle{\EHT}}$) is given by the 
grammar below, where the formulas outside $\HT$ are in bold.
\begin{align*}
\phi & \bnf p \mid \bot \mid \phi \land \phi \mid \phi \lor \phi \mid
\phi \intimp \phi \mid \boldsymbol{\K} \phi \mid 
\boldsymbol{\Khat} \phi \,.  ~~~~~~~~ (\text{for } p \in \Prop)
\end{align*}
As usual, the derived formulas $\intnot \phi$, $\top$, and $\phi \leftrightarrow \psi$ 
respectively abbreviate $\phi \intimp \bot$,
$\bot \intimp \bot$, and $(\phi \intimp \psi) \land (\psi \intimp \phi)$.
A theory is a finite set of formulas.
An $\EASP$-program $\epispec$ is translated to the corresponding
$\EHT$-theory $\trOf{\epispec}$ via a map $\trOf{(.)}$: given
$\Sigma =  \set{\mathtt{r_1}, \mathtt{r_2}}$ s.t.\ $ \mathtt{r_1}=
p \lor \classnot q \leftarrow \Khat\, r \land \notlp  s$ \ and \ 
$\mathtt{r_2}= q \leftarrow \lpnot \K\, p$,
\begin{align*}
\trOf{\Sigma} = \big(\ (\ \Khat\, r \land \intnot s \ ) \intimp ( \ p \lor \negof{q} \ )  \ \big)
~\land~ \big(\ \intnot \K\, p \intimp q \ \big)
~\land~ \intnot \big(\ q \land \negof{q} \ \big)\, .
\end{align*}
The literal $\classnot q$ is treated
as a new atom $\negof{q} \in \Prop$, and this
entails the formula $\intnot \big(q \land \negof{q})$ 
to be inserted into $\trOf{\Sigma}$ for consistency purposes.
Since it can be easily removed from a logic program with the addition of a constraint 
$\bot \leftarrow q \land \negof{q}$ as above, the rest of the paper disregards strong negation $\classnot$ for simplicity.

As already mentioned in Ex.\,\ref{ex: motovating ex 2 for EASP}, the $\Khat$ operator 
is syntactically different from $\M \in \lang{\scriptscriptstyle{\ES}}$. This is
justified by the fact that $\M$ is derived as $\lpnot \K\, \notlp$ in $\ES$ and so
translated into $\EHT$ as $\intnot \K \intnot$ where $\intnot$ refers to $\EHT$-negation. 
Because $\intnot \K \intnot \phi$, $\intnot \intnot \Khat \phi$, and
$\Khat \intnot \intnot \phi$ are
all equivalent in $\EHT$, the $\M$ operator is expected to coincide with
$\notlp \notlp \Khat$ and $\Khat \notlp \notlp$ in a possible extension of 
$\EASP$-programs
to propositional theories, which will be shortly discussed in the next section.
Notice that the difference between $\M\, p$ and $\Khat\, p$ in $\EASP$ resembles
that of $\notlp \notlp p$ and $p$ in $\ASP$. As a result,  in an extended language,
we expect $\M p$ not to have a world-view,
whereas $\set{\emptyset, \set p}$ is one easily-understandable 
world-view for $\Khat p$.

An $\EHT$-model $\tuple{\there, \weak}$ is a refinement of 
a classical $\sfive$-model $\there$
in which valuations $T \in \there$ are replaced by HT-models $(\weak(T), T)$ w.r.t.\ a
subset function $\weak {\suchthat} \there \rightarrow 2^{\Prop}$,
assigning to each $T \in \there$ one of its subsets, i.e., $\weak(T)\subseteq T$. 
Thus, the explicit representation of $\tuple{\there, \weak}$
is given by $\big\{\big(\weak(T),T\big)\big\}_{\scriptscriptstyle{T \in \there}}$.
Satisfaction of a formula $\phi \in \lang{\scriptscriptstyle{\EHT}}$ is defined recursively
w.r.t.\ to the following truth conditions:
$$\begin{array}{lll}
\tuple{\there, \weak},T \Ehtmodels p   & \text{ if } &  p \in \weak(T);
\\
\tuple{\there, \weak},T \Ehtmodels \phi \intimp \psi   & \text{ if } &
\big ( \tuple{\there, \weak},T \notEhtmodels \phi \text{ \, or \ } \tuple{\there, \weak},T  \Ehtmodels \psi \big )
\text{ \ and }
\\
 &&
\big ( \tuple{\there, id},T  \notEhtmodels \phi  \text{ \ or \ }  \tuple{\there, id},T  \Ehtmodels \psi \big );
\\
\tuple{\there, \weak},T \Ehtmodels \K \phi & \text{ if } &
\tuple{\there, \weak},T' \Ehtmodels \phi \text{ \, for every } T' \in \there;
\\
\tuple{\there, \weak},T \Ehtmodels \Khat \phi  & \text{ if } &
\tuple{\there, \weak},T' \Ehtmodels \phi \text{ \, for some } T' \in \there;
\end{array}$$
where $id$ denotes the identity function.
The truth conditions of $\bot$, $\land$ and $\lor$ are standard. The $\EHT$-model
$\tuple{\mathcal T,id} =\set{(T,T)}_{\scriptscriptstyle{T\in\mathcal T}}$ is called \emph{total} and identical to the classical $\sfive$-model $\mathcal T$.
Then, $\mathcal T$ is an \emph{equilibrium $\sfive$-model w.r.t.\ truth}, 
or originally
an \emph{epistemic equilibrium model} ($\EEM$)
of  $\phi \in \lang{\scriptscriptstyle{\EHT}}$
if $\mathcal T$ is a classical $\sfive$-model of $\phi$, 
and the following $\texttt{t}$-minimality condition
(referred to as \emph{$\tfunctional$-minimality}), viz.\ 
\begin{align} \label{truth minimality 1}
\text{for every possible subset function } \weak \text{ on } \mathcal T \text{ with }
\weak \neq id,
\text{ there is } T \in \mathcal T \text{ s.t.\ } 
\tuple{\there, \weak}, T \notEhtmodels \phi 
\end{align}
holds. $\ESyirmia$ further applies a knowledge-minimality
(\texttt{k}-minimality) criterion (\cite{SuAI20}, p.\,12), 
simultaneously functioning two different conditions, 
upon $\EEMs$ to determine its world-views, originally referred to as 
autoepistemic equilibrium models ($\AEEMs$). The inspiration comes from
autoepistemic logic and the logic of all-that-I-know, and the selection
process is carried out by mutual comparison of $\EEMs$
according to set inclusion and a formula-indexed
preorder. Note that applying the same criterion upon 
$\EASP$'s $\tfunctional$-minimal models to select world-views, we can
search for a relationship between two formalisms. 

%---------------------------------------------------------------------------------------------------------%
\subsection{Cabalar et al.'s approach {\normalfont (\texorpdfstring{$\ESyirmib$}{ESyirmib}):} 
founded autoepistemic equilibrium logic \normalfont (\texorpdfstring{$\FAEEL$}{FAEEL})}
\label{subsec:EEL.CFF}
%---------------------------------------------------------------------------------------------------------%
Cabalar et al.\ \cite{Cabalar20} define $\EHT$ on a $\Khat$-free
fragment of $\lang{\scriptscriptstyle{\EHT}}$. 
The authors acknowledge that the relation of 
a second operator ($\M$ vs.\ $\Khat$) 
to $\K$ is under debate, and so they leave its study for future work. 
Even though not in terms of meaning, the inclusion of $\Khat$ into $\ESyirmib$
is methodologically straightforward. Therefore, we here follow the same language 
$\lang{\scriptscriptstyle{\EHT}}$ for $\ESyirmib$ as well in terms of harmony. 
Moreover,
$\ESyirmib$ partially contains $\Khat$ when considered in its original language 
since $\intnot\Khat$ and $\Khat\intnot$ are $\EHT$-equivalent respectively to $\K \intnot$ and $\intnot\K$. As a derived formula, 
$\M$ is also included by default in all existing $\EEL$s
in the form of $\intnot \K \intnot$. Unlike in $\ESyirmia$ 
where $\K$ represents \emph{knowledge}, in this context,
$\K \phi$ reads \emph{$\phi$ 
is one of the agent's beliefs}. 

In $\ESyirmib$, an $\EHT$-model $\tuple{\there, \weakrelational}$ is defined  
w.r.t.\ a serial subset relation (multi-valued subset function) 
$\weakrelational$,
relating each $T \in \there$ to at least one element from $2^T$, i.e., to some subsets of $T$. Thus, 
a serial subset relation $\weakrelational$ and an $\sfive$-model $\mathcal T$ give rise to the $\EHT$-model
$\weakrelational =\set{(H,T) {\suchthat} T \weakrelational H}_{\scriptscriptstyle{T \in \mathcal T}}$. To illustrate the functional vs.\ relational
nature of the formalisms $\ESyirmia$ and $\ESyirmib$, 
take the $\sfive$-model $\mathcal T=\set{T}$ where $T=\set{p,q}$. Depending on the subset function $\weak$ on $\mathcal T$, we can only form the $\EHT$-models 
$\bigset{ (\emptyset , T)}$,  $\bigset{(\set p, T)}$,
$\bigset{(\set q , T)}$, and $\bigset{(T, T)}$ in $\ESyirmia$ as
we are restricted to choose a unique subset $H=\weak(T)$ and so build a unique $\HT$-model
$(H,T)$ for each $T \in \mathcal T$. However,
in $\EHTyirmib$, we can obtain the additional 
$\EHT$-models
\begin{align*}
\bigset{(\set p , T) , (\set q, T)}, \hspace{0.8em} 
\bigset{ (\emptyset, T) , (\set p , T), (\set q, T)}, \hspace{0.8em} 
\bigset{ (\emptyset , T) , (T, T)}, \hspace{0.8em} 
\bigset{ (\emptyset, T) , (\set p , T), (T, T)},  \text{ etc.}
\end{align*}
since as many subset as desired can be chosen for each
$T\in \mathcal T$, keeping in mind that $\weakrelational$ is serial.

While the truth conditions are the same, to avoid possible confusion, 
we recall that $\tuple{\there, \weak}, T \Ehtmodels \phi$ means 
$\set{(H,T) : H =\weak(T)}_{\scriptscriptstyle{T \in \mathcal T}},(H,T) \Ehtmodels \phi$ 
in $\ESyirmia$, 
but here $\weakrelational(T)$ may refer to more than 
one subset as $\weakrelational$ is multi-valued.
Thus, we prefer an explicit notation
$\set{(H,T) {\suchthat} T \weakrelational H}_{\scriptscriptstyle{T \in \mathcal T}}, (H,T) \Ehtmodels \phi$ to be precise.

An \emph{epistemic equilibrium model} ($\EEM$) of  $\phi \in \lang{\scriptscriptstyle{\EHT}}$ 
is then defined as its classical $\sfive$-model $\mathcal T$ satisfying a 
\emph{$\trelational$-minimality} condition: for every multi-valued subset function 
$\weakrelational$ on $\mathcal T$ s.t.\ $\weakrelational \neq id$,
\begin{align} \label{t-minimality 2}
\text{ there is an $\HT$-model } (H,T) \text{ s.t. } T \weakrelational H
\text{ and } 
\set{(H,T) {\suchthat} T \weakrelational H}_{\scriptscriptstyle{T \in \mathcal T}}, (H,T) 
\notEhtmodels \phi.
\end{align}
Once $\EEMs$ are produced, the next step is to apply Schwarz's \cite{Schwarz92} 
minimal model reasoning\footnote{Schwarz has proved that autoepistemic logic under stable expansions and $\KD$ under minimal models coincide.} for nonmonotonic $\KD$ to select world-views of $\ESyirmib$ from among $\EEMs$. The operator $\K$ obtains 
its meaning from this approach
because in autoepistemic logic, the epistemic operator $\K$ characterises the \emph{self-belief} of a rational agent. To weaken a $\trelational$-minimal $\sfive$-model ($\EEM$) w.r.t\ belief,
$\ESyirmib$ needs to generalise 
$\EEMs$ to $\KD$-model structures because minimality w.r.t.\ belief ({\texttt{b}}) is tested in nonmonotonic $\KD$ by examining whether an $\sfive$-model
has a \emph{preferred} model extension in $\KD$. To check stability w.r.t.\ belief in $\ESyirmib$,
we add a new valuation $I$ into a (candidate) 
$\EEM$ $\mathcal T$ s.t.\ $I \not\in \mathcal T$ 
and design the resulting $\KD$ model $\mathcal T' = \mathcal T \uplus \set I$ in a way that $I$ is not accessible by any point in $\mathcal T'$ while any point in $\mathcal T$ can be accessed by every point in $\mathcal T'$.
Thus every point in $\mathcal T'$, including $I$, 
uses the same belief that is determined by $\mathcal T$. Formally,
$\mathcal T'$ is preferred over $\mathcal T$, and  $\mathcal T'$ is a $\trelational$-minimal $\KD$-model of $\phi$ if the following conditions
\begin{align} \label{KD45 weakening}
\text{(i) }~ \mathcal T \uplus \set{I} \ , \ I \ \KDmodels \phi \text{ \, \, \, and \, \, \, } 
\text{(ii) }~ \mathcal T \uplus \set{(\weak(I), I)} \ , \ (\weak(I), I) \ \notEhtmodels \phi
\end{align}
respectively hold.
When (\ref{KD45 weakening}) holds for a candidate $\EEM$ $\mathcal T$, 
this means that
$\mathcal T$ is not stable (or at equilibrium) w.r.t.\ 
belief and fails to be an $\AEEM$ of $\phi$ in $\ESyirmib$.
Notice that the condition (\ref{KD45 weakening}).(ii) does not require that 
the points of $\mathcal T$ be weakened w.r.t.\ truth: 
as $\mathcal T$ is an EEM of $\phi$, by definition, 
any weakeaning of $\mathcal T$ results in the formula $\phi$ being refuted
at some point of $\mathcal T$. Also note that due to the $\KD$-model structure, 
we weaken $I$ w.r.t.\ truth in $\ESyirmib$
simply by using the subset function $\weak$ in (\ref{KD45 weakening}) as $\weakrelational$ and $\weak$
provide identical models. We  do not 
reformulate above the details of the method in its original notation 
as our aim here is to give
a brief overview to the reader. However, for our purposes, it is worth mentioning that 
this $\texttt{b}$-minimality approach can be easily adapted to $\trelational$-minimal 
models of $\EASP$ as formalised below. 
\begin{definition} [\textbf{stable $\sfive$-models of $\EASP$ w.r.t.\ truth and belief}]
\normalfont Let $\mathcal T$ be a nonempty collection of valuations, and
let $\epispec$ be an $\EASP$-program. Then, $\mathcal T$ is a 
\emph{stable $\sfive$-model} of $\epispec$ w.r.t.\ truth and belief 
if for every $T \in \mathcal T$, we have
$\mathcal T, T \starmodelsrelational \epispec^{\tuple{\mathcal T, T}}$  \ and \
for every $I \in 2^{\Prop} \setminus \mathcal T$,
\begin{align} \label{defn:stable sfive models}
\mathcal T \uplus \set I \ , I \not \KDmodels \epispec^{\tuple{\mathcal T, I}} \text{ ~or~ }
\mathcal T \uplus  \set{(\weak(I), I)}\ , (\weak(I),I) \KDmodels \epispec^{\tuple{\mathcal T, I}}
\text{ for some subset map }  \weak  \text{ s.t.\ } \weak(I) \subset I. 
\end{align}
\end{definition}

The condition (\ref{defn:stable sfive models}) states that $\mathcal T$ has no $\trelational$-minimal preferred model in $\KD$. 
This definition will then allow us to search for a correspondence 
between the resulting formalism and $\ESyirmib$.

%---------------------------------------------------------------------------------------------------------------%
\subsection{Su's approach {\normalfont (\texorpdfstring{$\ESyirmibir$}{ESyirmibir}):} 
reflexive autoepistemic equilibrium logic \normalfont (\texorpdfstring{$\RAEEL$}{RAEEL})}
\label{subsec:EEL.Su}
%----------------------------------------------------------------------------------------------------------------%
Su \cite{Su2021} then suggests applying the $\texttt{k}$-minimality
criterion of nonmonotonic $\SW$ \cite{Su17.reflexive} over $\EEMs$ of $\ESyirmia$ or $\ESyirmib$ to select $\AEEMs$ and proposes $\ESyirmibir$. Remember that the modal logic $\SW$ is just a reflexive\footnote{Schwarz \cite{Schwarz92} has proved that reflexive 
autoepistemic logic and nonmonotonic $\SW$ coincide under their specific semantics.} closure of 
$\KD$ where $\K$ represents \emph{knowledge}. 
Our underlying intuition is simply because  
the formulas $\K\,p$ and $\Khat\,p$ have respectively the unique AEEMs 
$\set{\set p}$ and $\set{\emptyset, \set p}$ in $\ESyirmibir$, regardless of the 
$\texttt{t}$-minimality technique chosen, $\trelational$ vs.\ $\tfunctional$. 
While $\Khat\,p$ has the same AEEM, $\K\,p$ has no AEEM
in $\ESyirmib$. In an extended language, $\ESdoksandort$ cannot provide any 
world views for these formulas, and a slightly modified version
$\ESonbir$ \cite{gelfond2011new} cannot produce
a reasonable model $\set{\set p}$ for $\K\,p \lor q$. These results reinforce the counter-arguments 
provided in Ex.\,\ref{ex: motovating ex 1 for EASP}-\ref{ex: motovating ex 2 for EASP}  
towards their reduct definitions, attacking positive subjective literals. If an atom $p$ can be derived in all
stable models of an $\ASP$-program, then the query $p?$ is answered as \emph{true}. 
Does it provide an enough justification for the derivation of $\K\,p$? 
While $p$ has a unique world view $\set{\set p}$, 
why does a stronger expression $\K\,p$ lack a world-view? Such questions go on...
Although it is unclear what researchers intend to capture with $\K$, 
the above-mentioned
$\EEL$s, especially $\ESyirmia$ and $\ESyirmibir$ with their well-studied minimality tools, 
are strong formalisms, and 
in our opinion, they both can serve
with their different functionalities (especially towards constraints) for the encoding of different problems.

All existing $\EEL$s in the literature
employ a twofold world-view computation process. The method
is first to compute $\texttt{t}$-minimal models of a program, upon which a $\texttt{k}$-minimality criterion is applied.
In $\ESdoksandort$,
there is no such clear-cut distinction between truth and knowledge minimality conditions; 
instead, they are given intertwined with each other. The follow-up $\ES$-formalisms are 
mostly focused on reduct without modifications in the minimality. 
This makes it difficult to understand
the relationships between $\ES$-formalisms proposed in the logic programming domain 
and the purely logical domain of $\EL$. However, there are some work in the literature, 
revealing similarities between existing $\ES$-formalisms. For instance,
Wang\&Zhang \cite{WangZ05nested}) have embedded $\ESdoksandort$ into an $\EEL$
they designed;
Cabalar et al.\ \cite{Cabalar20} have proved that
AEEMs of $\ESyirmib$ and founded world-views of $\ESdoksandort$ coincide 
under a foundedness property they proposed. We tackle this research topic
in reverse direction by following Ferraris' work, which captures equilibrium models as stable models. 
To achieve this, we propose a versatile and comprehensive framework called 
$\EASP$ that can evolve into various $\EEL$s, incorporating their $\texttt{k}$-minimality conditions. 
Moreover, compared to related work, it is evident how $\EASP$ accommodates existing 
$\EEL$s through the traditional nature of $\EASP$. 
The next section clarifies how we accomplish this in a unifying framework.

%------------------------------------------------------------------------------------------------------------%
\section{Correspondence between \texorpdfstring{$\EASP$}{EASP} 
and \texorpdfstring{$\EEL$}{EEL} }
\label{sec:correspondence}
%------------------------------------------------------------------------------------------------------------%
This section first generalises Ferraris' lemma, presented in (\cite{Ferraris05a}, p.\,3), 
to $\sfive$, $\KD$ and $\SW$-models.
\begin{lem}\label{lem:Ferraris}
Given $I=\set{1, \ldots , n}$,
let $\mathcal T = \set{T_i}_{i \in I} =\set{T_i \suchthat i \in I}$ be an
$\sfive$-model, and
let $\weak\suchthat \mathcal T \rightarrow 2^{\Prop} $ be a subset function s.t.\ $\weak(T_i)=H_i \subseteq T_i$ for every $i \in I$.
For an $\EASP$-program $\epispec$, 
\begin{align*}
\set{H_1, \ldots, H_n}~,~H_j~\sfivemodels \epispec^{\tuple{\mathcal T, T_j}} 
\text{ \  \ iff \,  \ }
\set{(H_i,T_i) : i \in I}~,~(H_j,T_j)~\Ehtmodels \epispec^*, 
\text{ \ \ \ for every $j\in I$.}
\end{align*}
\end{lem}
The lemma is proven by structural induction. 
As $H_i = H_j$ is possible for some $i,j \in I$, 
we consider $\set{H_i}_{i\in I}$ as a multiset and employ the traditional 
reduct introduced in Def.\,\ref{defn:epis.reduct}. Under this general result, we can clearly see
how EELs appear in the logic programming domain and vice versa. 

We begin with $\EEMs$ of $\ESyirmia$:
for an $\EASP$-program $\epispec$, let 
$\mathcal T=\set{T_i}_{i \in I}=\set{T_1, \ldots, T_n}$ be an 
$\EEM$ of $\epispec^*$. By definition of $\EEM$ in $\ESyirmia$ 
($\ref{truth minimality 1}$), we have
(1) $\mathcal T,T_i \sfivemodels \epispec^*$ for every $i \in I$ and
(2) for every non-identity subset function $\weak$ on $\mathcal T$
s.t.\ $\weak(T_i)=H_i$ for each $i$, there is $k\in I$ s.t.\ $\tuple{\mathcal T,\weak}, T_k \notEhtmodels \epispec^*$. The model $\tuple{\mathcal T,\weak}$ gives rise to the $\EHT$-model 
$\set{(H_i,T_i)}^{n}_{\!i=1}=\set{(H_1,T_1),\ldots, (H_n,T_n)}$, and so, 
$\set{(H_i,T_i)}^{n}_{\!i=1}, (H_k,T_k) \notEhtmodels \epispec^*$.
First let $\weak = id$ in Lemma \ref{lem:Ferraris}, then $H_i=T_i$  for each $i$.
Recall that $\tuple{\mathcal T, id}$ refers to the classical $\sfive$-model $\mathcal T$. The condition (1) so implies 
$\set{T_1 \ldots T_n}, T_j \sfivemodels \epispec^{\tuple{\mathcal T,T_j}}$,
for every $j\in I$. 
Again by Lemma \ref{lem:Ferraris}, the condition (2) refers to a more relaxed 
$\tfunctional$-minimality criterion 
not discussed in Sect.\,\ref{sec:EpisASP}, saying that
\doublequotes{for every subset function $\weak$ with $\weak \neq id$, 
there is $k \in I$ s.t.\
$\set{\weak(T_1), \ldots, \weak(T_n)}, \weak(T_k) \not\sfivemodels \epispec^{\tuple{\mathcal T, T_k}}$}.
To sum up, we have:
\begin{align}\label{correspondence:EEMs in EASP}
&\mathcal T, T \sfivemodels \epispec^{\tuple{\mathcal T,T}} \text{ for every } T \in \mathcal T 
\text{\ \ \ and  \ \ \ \ } 
\\
&\text{for every subset function } \weak \neq id, \text{ there is } T' \in \mathcal T 
\text{ s.t.\ }
\set{\weak(T)}_{T\in \mathcal T}, \weak(T') \notsfivemodels \epispec^{\tuple{\mathcal T, T'}}. 
\nonumber
\end{align}

Since $\EHT$-models of $\ESyirmib$ are formed in a relational structure, 
first we should refine Lemma \ref{lem:Ferraris}.
\begin{lem}\label{lem:Ferraris2}
Let $\epispec$ be an $\EASP$-program. Let $\mathcal T$ be an $\sfive$-model, and
let $\weakrelational$ be a multi-valued
subset function on $\mathcal T$ s.t.\ $\weakrelational =
\set{(H,T) : T \weakrelational H}_{\scriptscriptstyle{T \in \mathcal T}}$. 
For every $T \in \mathcal T$, let $H$ be s.t.\ 
$T\weakrelational H$. Then, we have: 
\begin{align*}
\set{H : T\weakrelational H}_{\scriptscriptstyle{T \in \mathcal T}}~,~H~\sfivemodels \epispec^{\tuple{\mathcal T, T}} \text{ \  \ iff \,  \ }
\set{(H,T) : T \weakrelational H}_{\scriptscriptstyle{T\in \mathcal T}}~,~(H,T)~\Ehtmodels \epispec^*.
\end{align*}
\end{lem}

Pursuing a similar proof, we can also capture $\EEMs$ of $\ESyirmib$ in $\EASP$.
While the line (\ref{correspondence:EEMs in EASP}) remains the same, we again obtain a more-relaxed 
$\trelational$-minimality condition compared to one proposed in Sect.\,\ref{sec:EpisASP}:
\begin{align}\label{correspondence:foundedEEMs in EASP}
&\text{for every multi-valued subset function } \weakrelational \text{ s.t.\ } \neq id,
\nonumber 
\\
&\set{H : T\weakrelational H}_{\scriptscriptstyle{T\in \mathcal T}}~,~H'~\notsfivemodels \epispec^{\tuple{\mathcal T, T'}}
\text{ for some } T' \in \mathcal T 
\text{ and for some } H' \text{ s.t.\ } T' \weakrelational H'. 
\end{align}

The extensions of Lemma \ref{lem:Ferraris}-\ref{lem:Ferraris2} to
$\KD$ and $\SW$-model structures and 
reflecting generalised $\EEMs$ in such weaker model structures
to $\EASP$ are straightforward. 
We now perform the same task in the opposite direction and
embed Def.\,\ref{defn:epis.reduct} into $\EEL$
domain. Using Lemma \ref{lem:Ferraris}, $\mathcal T=\set{T_i}_{i\in I}$ is a $\tfunctional$-minimal $\sfive$-model of $\epispec$ \ iff \
$\mathcal T, T_i \sfivemodels \epispec^*$ for every $i \in I$  and for every $j \in I$,
we have
\begin{align}\label{t.min models in EEL}
\text{for every subset map } \weak \text{ s.t.} \restr{\weak}{\mathcal T\setminus \set{T_j}}=id 
\text{ and } \weak(T_j)\subset T_j,\,\,\,
\set{(\weak(T_i),T_i)}_{i}, (\weak(T_j),T_j)\notEhtmodels \epispec^*. 
\end{align}

We leave it to the reader to generalise this result to 
$\trelational$-minimal $\sfive$-models of $\EASP$
by Lemma \ref{lem:Ferraris2}.

Through the same approach, we try to analyse $\ESdoksandort$-semantics:
let $\mathcal T=\set{T_1, \ldots, T_n}$ be a 
world-view of an $\ES$-program $\epispec$. By definition, 
$\mathcal T$ is the maximal set w.r.t.\
subset relation $\subseteq$ satisfying 
(1) $\mathcal T \models \epispec^{\mathcal T}$ and 
(2) $(\mathcal T\setminus\set{T}) \cup\set{H}, H \not\models \epispec^{\mathcal T}$ 
for every $H \subset T$, for every $T \in \mathcal T$. 
Notice that $\epispec^{\mathcal T}=\epispec^{\tuple{\mathcal T,T_i}}$ for every $i \in I$ as $\ESdoksandort$-reduct eliminates only extended subjective literals. This definition, except maximality condition, coincides with $\tfunctional$-minimal $\sfive$-model definition of $\EASP$, and so with (\ref{t.min models in EEL}) by Lemma \ref{lem:Ferraris}. However,
$\set{\set p}$ is a world-view of $p \leftarrow \K p$, but not a $\tfunctional$-minimal $\sfive$-model of $\epispec$.
For some reasons, we cannot apply Ferraris' generalised lemma 
(i.e., Lemma \ref{lem:Ferraris}) to $\ESdoksandort$-semantics. 

%--------------------------------------------------------------------------------------------------------------%
\section{Conclusion}
\label{sec:conclusion}
%--------------------------------------------------------------------------------------------------------------%
In this paper, we first discuss the problems that arise under Gelfond's original $\ESdoksandort$-semantics, aiming to
shed light on the underlying reasons for these issues. 
We also briefly overview the follow-up semantics, that were 
primarily proposed to address the limitations of $\ESdoksandort$. 
Next, we introduce a flexible and robust 
framework for epistemic logic programs called $\EASP$, which 
already accommodates Su's traditional $\tfunctional$-minimal $\sfive$-models, 
as studied in \cite{Su19jelia}, and 
their novel variations known as $\trelational$-minimal $\sfive$-models. 

We recognise that all existing epistemic equilibrium logics ($\EEL$s) in the literature 
share a two-step world-view computation process. This motivates us to explore their similarities and beyond
within the $\EASP$ context. To this end,
we generalise Ferraris' lemma (see \cite{Ferraris05a}, p.\,3), which establishes
a correlation between stable models and equilibrium models, to the epistemic case.
We then examine how these $\EEL$s are reflected within the $\EASP$ framework. 
This approach also allows us to investigate whether 
different $\tfunctional$ ($\trelational$) minimality methods, such as
those presented in \cite{Su19jelia} and \cite{SuAI20}, produce the same results 
when considered at least within 
the current $\EASP$ language fragment. It is worth noting that the technique in
\cite{Su19jelia} is 
slightly easier than that in \cite{SuAI20}, which
raises an immediate research question for future studies. Furthermore,
Ferraris' generalised lemmas lead to the strong equivalence 
characterisations of
$\EASP$-programs through the logical equivalences of their translations in $\EHT$, 
akin to Lifschitz et al.'s finding \cite{ASP-strong.equivalence} 
in regular $\ASP$. 
Finally, future work will also involve a more detailed investigation of how
Gelfond's $\ESdoksandort$-semantics can be reflected into the $\EEL$ domain,
following a similar approach as discussed in this paper. This study will help us better identify
the problems of $\ESdoksandort$, as well as its 
possible similarities with other $\ES$-semantic approaches originally proposed
in the $\EEL$ domain.

\bibliographystyle{eptcs}
\bibliography{iclp24_paper3}
\end{document}